\documentclass[conference, letterpaper]{IEEEtran}

\ifCLASSINFOpdf
\else
\fi
\usepackage{subcaption}

\ifCLASSINFOpdf
   \usepackage[pdftex]{graphicx}
\else
\fi

\usepackage[cmex10]{amsmath}
\usepackage{color}
\usepackage{fancyhdr}

\renewcommand{\thispagestyle}[2]{} 

\fancypagestyle{plain}{
        \fancyhead{}
        \fancyhead[C]{first page center header}
        \fancyfoot{}
        \fancyfoot[C]{first page center footer}
}
\begin{document}

%
\title{Dynamic Control of Explore/Exploit Trade-Off In Bayesian Optimization}


\author{\IEEEauthorblockN{Dipti Jasrasaria and Edward O. Pyzer-Knapp}
\IEEEauthorblockA{IBM Research\\
Hartree Centre\\
Sci-Tech Daresbury\\
Email: epyzerk3@uk.ibm.com}}


%


\maketitle

\begin{abstract}
Bayesian optimization offers the possibility of optimizing black-box operations not accessible through traditional techniques. The success of Bayesian optimization methods such as Expected Improvement (EI) are significantly affected by the degree of trade-off between exploration and exploitation.  Too much exploration can lead to inefficient optimization protocols, whilst too much exploitation leaves the protocol open to strong initial biases, and a high chance of getting stuck in a local minimum.  Typically, a constant margin is used to control this trade-off, which results in yet another hyper-parameter to be optimized. We propose contextual improvement as a simple, yet effective heuristic to counter this - achieving a one-shot optimization strategy.  Our proposed heuristic can be swiftly calculated and improves both the speed and robustness of discovery of optimal solutions.  We demonstrate its effectiveness on both synthetic and real world problems and explore the unaccounted for uncertainty in the pre-determination of search hyperparameters controlling explore-exploit trade-off.
\end{abstract}



%
\IEEEpeerreviewmaketitle

\section{Introduction}

Many important real-world global optimization problems are so-called `black-box' functions - that is to say that it is impossible either mathematically, or practically, to access the object of the optimization analytically - instead we are limited to querying the function at some point $\textbf{x}$ and getting a (potentially noisy) answer in return. Some typical examples of black-box situations are the optimization of machine-learning model hyper-parameters \cite{snoek_practical_2012,bergstra_hyperopt:_2015}, or in experimental design of new products or processes. \cite{lisicki_optimal_2016} 

One popular framework for optimization of black-box functions is Bayesian optimization. \cite{snoek_practical_2012,brochu_tutorial_2010,mockus_bayesian_1974,shahriari_taking_2016,mockus_bayesian_1982}

In this framework, a Bayesian model (typically a Gaussian process\cite{rasmussen_gaussian_2006,snoek_practical_2012}, although other models have been successfully used\cite{snoek_scalable_2015}) based on known responses of the black-box function is used as an ersatz, providing closed form access to the marginal means and variances.  The optimization is then performed upon this 'response surface' in place of the true surface. The model's prior distribution is refined sequentially as new data is gathered  by conditioning it upon the acquired data, with the resulting posterior distribution then being sampled to determine the next point(s) to acquire.  In this way, all else being equal, the accuracy of the response surface should start to increasingly resemble the true surface.  This is in fact dependent upon some of the choices made in the construction of the Bayesian model; and it is worth noting that a poor initial construction of the prior, through for instance an inappropriate kernel choice, will lead to a poor optimization protocol. 

Since Bayesian optimization does not have analytical access properties traditionally used in optimization, such as the gradients, it relies upon an acquisition function being defined for determining which points to select.  This acquisition function takes the model means and variances derived from the posterior distribution and translates them into a measure of the predicted utility of acquiring a point.  At each iteration of Bayesian optimization, the acquisition function is maximized, with those data points corresponding to maximal acquisition being selected for sampling. 

Bayesian optimization has particular utility when the function to be optimized is expensive, and thus the number of iterations the optimizer can perform is low.  It also has utility as a 'fixed-resource optimizer' since - unlike traditional optimization methods - it is possible to set a strict bound on resources consumed without destroying convergence criteria.  Indeed, in abstract, the Bayesian optimization protocol of observe, hypothesize, validate is much closer in spirit to the scientific method than other optimization procedures. 

\subsection{Acquisition Functions}
A good choice of acquisition function is critical for the success of Bayesian optimization, although it is often not clear \textit{a priori} which strategy is best suited for the task.  Typical acquisition strategies fall into one of two types - improvement based strategies, and information based strategies.  An improvement based strategy is analogous to the traditional optimization task in that it seeks to locate the global minimum/maximum as quickly as possible.  An information based strategy is aimed at making the response surface as close to the real function as quickly as possible through the efficient selection of representative data. Information based strategies are strictly exploratory and thus we focus our attention on improvement based strategies for the duration of this paper.

In general, we can define the improvement, $\gamma$, provided by a given data-point, $x$, as 

\begin{equation}
\gamma(x)  = \frac{\mu(x) - f^{*}}{\sigma(x)}
\label{eq:improvement_max}
\end{equation}
for maximization, where $f^{*}$ is the best target value observed so far, $\mu(x)$ is the predicted means supplied through the Bayesian model, and $\sigma^{2}$ are their corresponding variances. 

Two typically used acquisition functions are the \textit{Probability of Improvement} (PI) \cite{kushner_new_1964} and the \textit{Expected Improvement} (EI).\cite{mockus_bayesian_1974} In PI, the probability that sampling a given data-point, $x$, improves over the current best observation is maximized:
\begin{equation}
PI(x) = \Phi(\gamma(x))
\label{eq:pi}
\end{equation}

where $\Phi$ is the CDF of the standard normal distribution.

One problem with the approach taken in PI is that it will, by its nature, prefer a point with a small but certain improvement over one which offers a far greater improvement, but at a slightly higher risk.  In order to combat this effect, Mockus proposed the EI acquisition function.\cite{mockus_bayesian_1974}  A perfect acquisition function would minimize the expected deviation from the true optimum, $f(x^{*})$, however since that is not known (why else would we be performing optimization?) EI proposes maximizing the expected improvement over the current best known point:
\begin{equation}
EI(x) = \mu(x) - f^{*}\Phi(\gamma) + \sigma(x) \phi(\gamma)
\label{eq:ei}
\end{equation}
where $\phi$ denotes the PDF of the standard normal distribution.

By maximizing the expectation in this way, EI is able to more efficiently weigh the risk-reward balance of acquiring a data point, as it considers not just the probability that a data point offers an improvement over the current best, but also how large that improvement will be.  Thus a larger, but more uncertain, reward can be preferred to a small but high-probability reward (which would have been selected using PI). 

EI has been shown to have strong theoretical guarantees \cite{vazquez_convergence_2010}and empirical effectiveness \cite{snoek_practical_2012} and so we use it throughout this study as the baseline.  

\section{Contextual Improvement}

\subsection{Exploration vs. Exploitation Trade-Off}

As with any global optimization procedure, in Bayesian optimization there exists a tension between exploration (i.e. the acquisition of new knowledge) and exploitation (i.e. the use of existing knowledge to drive improvement).  Too much exploration will lead to an inefficient search, whilst too much exploitation will likely lead to local optimization - potentially missing completely a much higher value part of the information space. 

EI, in its naive setting, is known to be overly greedy as it focuses too much effort on the area in which it believes the optimum to be, without efficiently exploring additional areas of the parameter space which may turn out to be more optimal in the long-term.  The addition of margins to the improvement function in \textbf{ Equation \ref{eq:improvement_max}} allow for some tuning in this regard. \cite{jones_taxonomy_2001,lizotte_practical_2008}  A margin specifies a minimum amount of improvement over the current best point, and is integrated into \textbf{Equation \ref{eq:improvement_max}} as follows:

\begin{equation}
\gamma  = \frac{y_{pred} - f^{*} + \epsilon}{\sigma}
\label{eq:imp_eps_max}
\end{equation}
for maximization, where $\epsilon \geq 0$ represents the degree of exploration.  The higher $\epsilon$, the more exploratory.  This is due to the fact that high values of $\epsilon$ require greater inclusion of predicted variance into the acquisition function.  

\subsection{Definition of Contextual Improvement}

The use of modified acquisition functions such as  \textbf{Equation \ref{eq:imp_eps_max}} have one significant drawback. Through their use of a constant $\epsilon$ whose value is determined at the start of sampling, they now include an additional hyperparameter which itself needs tuning for optimized performance.  Indeed the choice of $\epsilon$ can be the defining feature for the performance of the search.  As Jones notes in his 2001 paper \cite{jones_taxonomy_2001}:

\textit{...the  difficulty  is  that [the optimization method] is extremely sensitive to the choice of the target.  If  the  desired  improvement  is  too  small,  the  search  will  be  highly local and will only move on to search globally after searching nearly exhaustively around the current best point.  On the other hand, if $\epsilon$ is  set  too  high,  the  search  will  be  excessively  global,  and  the algorithm will be slow to fine-tune any promising solutions}

Given that the scope of Bayesian optimization is for optimizing functions whose evaluations are expensive; this is clearly not desirable at all.  

In order to combat this, we propose a modification of the improvement which is implicitly tied to the underlying model, and thus changes dynamically as the optimization progresses - since the exploration / exploitation trade-off is now dependent upon the model's state at any point in time, we call this \textit{contextual improvement}, or $\chi$:

\begin{equation}
\chi = \frac{y_{pred} - f^{*} + c_{v}}{\sigma}
\label{eq:contextual_improvement}
\end{equation}
for maximization, where $c_{v}$ is the contextual variance for which can be written as:
\begin{equation}
c_{v} = \frac{\overline{\sigma^{2}}}{f^{*}}
\label{eq:coeff_of_var}
\end{equation}
where $\overline{\sigma^{2}}$ is the mean of the variances contained within the sampled posterior distribution and should be distinguished from $\sigma$ which is the individual variance of a prediction for a particular point in the posterior .

This is an intuitive setting for improvement, as exploration is preferred when, on average, the model has high uncertainty, and exploitation is preferred when the predicted uncertainty is low.   This can provide a regularization for the search, due to the effects an overly local search will have on the posterior variance.  The rationale for this is as follows: since the posterior variance can be written as 
\begin{equation}
\sigma^{2}(x_{*}) = K(X_{*},X_{*}) - K(X_{*}, X)K(X, X)^{-1} K(X, X_{*})
\label{eq:gp_var}
\end{equation}
 where $X_{*}$ represents a set of as yet unsampled data-points (i.e. part of the posterior rather than the prior), $K$ represents the kernel function and therefore  $K(X,X_{*})$  denotes the $n x n_{*}$
covariance matrix evaluated at all pairs of training $(X)$ and test
$X_{*}$ points, and similarly for $K(X,X)$, and $K(X_{*},X_{*})$, \cite{rasmussen_gaussian_2006} - we can see that the variance depends only upon the feature space.  If a search is overly local (i.e. stuck in a non-global minimum), it will produce a highly anisotropic variance distribution with small variances close to the local minima sampled, and larger variances elsewhere in the information space.  This results in a larger value for the standard deviation for the posterior variance, which in turn, through \textbf{Equation \ref{eq:contextual_improvement}}, forces greater sampling of the variance (equivalent to an increase in $\epsilon$).  Since the variance is low in the locally sampled area, the acquisition function is depressed here.  It is important to note here, the difference between this approach and an \textit{information centered approach}.  Due to the fact that \textbf{Equation \ref{eq:contextual_improvement} }works directly on the acquisition function, if there are no other areas with a high expectation of improvement (i.e. the local optimum is also predicted to be a strong global optimum beyond the range of variance) then that area will continue to be sampled - this is not the case in an information centered approach. 

When the acquisition function is optimized directly (using a global optimization technique such as DIRECT - DIviding RECTangles),\cite{noauthor_direct_nodate} the authors suggest providing a value for the distribution of the posterior variance required for \textbf{Equation \ref{eq:contextual_improvement}}, $\overline{\sigma^{2}}$, using a sampling method over the function bounds such as a low-discrepency sequence generation such as a Sobol or Halton sequence.  Alternatively, if the manifold is not suited to this type of exploration, an MCMC-type sampling method such as slice sampling \cite{neal_slice_2003,murray_slice_2010} will also produce satisfactory results, albeit at greater computational expense.


\section{Experiments}
\subsection{Definition of Success Metrics}

In order to separate the contribution of contextual improvement from other algorithmic contributions, we directly compare EI with traditional improvement, $\epsilon$-EI, with a value of $0.3$ (a common value for $\epsilon$) and EI using contextual improvement, which we will denote as adaptive EI (AEI).   Our metrics for success are twofold: firstly, we measure the performance of the search (i.e. which method finds, on average, the best value) - this is referred to in the results tables as \textbf{Mean} - and secondly we measure the robustness of the search (how much variance is there between repeat searches).  The robustness is measured as the difference between the 10th and 90th confidence intervals of the final sampling point (i.e. 50th) as calculated using a bootstrap. Thus, throughout this study robustness is referred in results tables as \textbf{$\Delta CI $}.  

\subsection{Experimental Details}

For all experiments, we utilize a Gaussian process with a squared-exponential kernel function with ARD using the implementation provided in the GPFlow package.\cite{matthews_gpflow:_2016}  We optimized the hyperparameters of the Gaussian process at each sampling point on the log-marginal likelihood with respect to the currently observed data-points.  The validity of the kernels was determined by testing for vanishing length-scales as this is typically observed when the kernel is miss-specified.   Each experiment was repeated 10 times, with confidence intervals being estimated using bootstrapping of the mean function.

\subsection{Optimization of Synthetic Functions}

One of the traditional ways of evaluating the effectiveness of Bayesian optimization strategies is to compare their performance on synthetic functions.  This has the advantage of the fact that these functions are very fast to evaluate, and the optima and bounds are well known.  Unfortunately these functions are not necessarily representative of real world problems, hence the inclusion of the other two categories.  We have chosen to evaluate three well-known benchmarking functions, the Branin-Hoo function (2D, minimization), the 6-humped camelback function (2D, minimization), and the 6-dimensional Hartmann function (6D, maximization).    

\subsection{Tuning of Machine Learning Algorithms}

A popular use for Bayesian optimization functions is for tuning the hyperparameters of other machine-learning algorithms. \cite{snoek_practical_2012,bergstra_hyperopt:_2015} Due to this fact, the lack of dependence of contextual improvement on pre-set scheduling hyperparameters is particularly important.  In order to test the effectiveness of contextual improvement for this task, we use it to determine optimal hyperparameters for a support vector machine for the abalone regression task.\cite{nash_population_1994}  In this context we have three hyperparameters to optimize -  $C$ (regularization parameter), $\epsilon$ (insensitive loss) for regression and $\gamma$ (RBF kernel function).  For the actual prediction process, we utilize the support vector regression function in scikit-learn.\cite{pedregosa_scikit-learn:_2011}  We also tune five hyperparameters of a 2 layer multi-layered perceptron to tackle the MNIST 10-class image classification problem\cite{lecun_mnist_2010} for handwritten digits. Here we tune the number of neurons in each layer,  the level of dropout \cite{srivastava_dropout:_2014}in each layer, and the learning rate for the stochastic gradient descent using the MLP implementation provided in the keras package,\cite{chollet_keras_2015} which was used in conjunction with TensorFlow.\cite{abadi_tensorflow:_2015}

\subsection{Experimental Design}

An obvious use for Bayesian optimization is in experimental design, where each evaluation can be expensive both in time and money, and the targets can be noisy.  For this experiment, we aim to design 2D aerofoils which optimize a lift to drag ratio as calculated using the JavaFoil program.\cite{hepperle_javafoil_nodate}  In order to specify the aerofoil design, we use the NACA 4-digit classification scheme, which denotes thickness relative to chord length, camber relative to chord length, the position of the camber along the chord length, as well as the angle of attack, thus resulting in a 4-dimensional optimization problem.  It is important to note that, due to the empirical treatment of the drag coefficient, unrealistically high values of the lift to drag ratio can be observed when using JavaFoil as the ground truth.  We chose to simply optimize the ground-truth function as calculated, but note the potential to apply a constraint in the optimization to account for this. \cite{gelbart_bayesian_2014}

\onecolumn
\begin{figure}[h]
 \begin{minipage}[0.7\textheight]{\textwidth}
\begin{subfigure}{.47\textwidth}
\includegraphics[width=\linewidth]{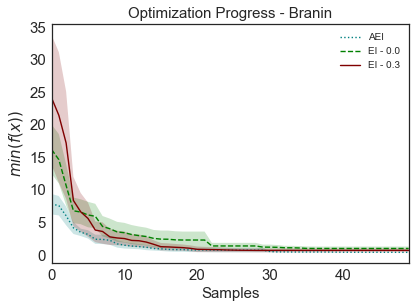}
\caption{•}
\end{subfigure}
\begin{subfigure}{.47\textwidth}
\includegraphics[width=\linewidth]{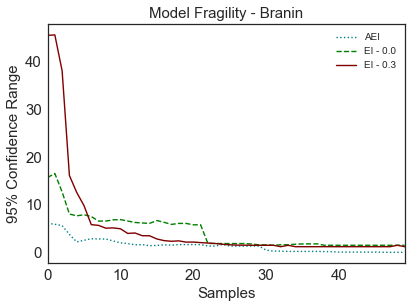}
\caption{•}
\end{subfigure}
\begin{subfigure}{.47\textwidth}
\includegraphics[width=\linewidth]{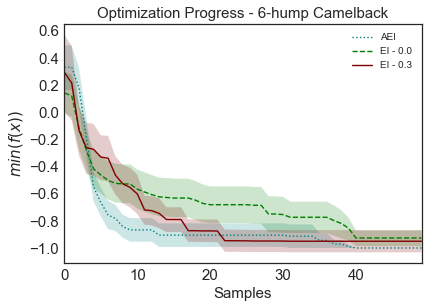}
\caption{•}
\end{subfigure}
\begin{subfigure}{.47\textwidth}
\includegraphics[width=\linewidth]{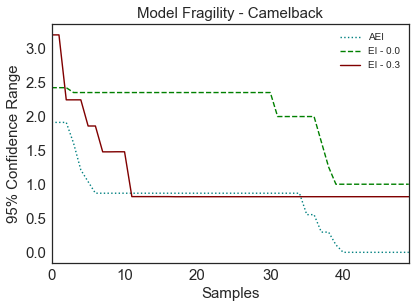}
\caption{•}
\end{subfigure}
\begin{subfigure}{.47\textwidth}
\includegraphics[width=\linewidth]{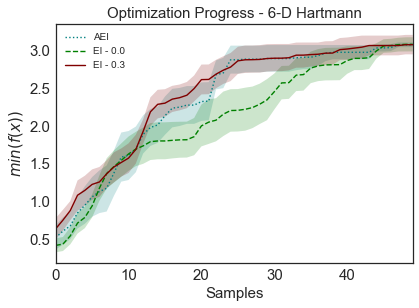}
\caption{•}
\end{subfigure}
\hspace{0.8cm}
\begin{subfigure}{.47\textwidth}
\includegraphics[width=\linewidth]{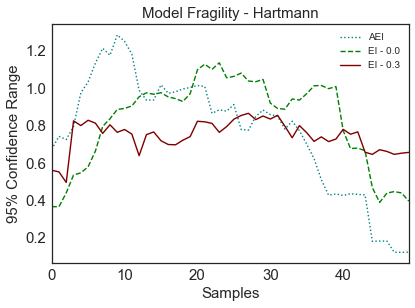}
\caption{•}
\end{subfigure}
\caption{Summary of the searches performed on synthetic functions.  (a), (c) and (e) show the evolution of the best sampled value for Branin,  camelback and 6-D Hartmann respectively, whilst (b), (d) and (f) show the evolving fragility of the model based upon the initial seed data.  Values are constructed by bootstrapping the mean over each equivalent sampling position as the search progresses.  If the performance of the model is strongly varying amongst the 10 trial runs performed then the value is large.  Since we are aiming at developing a robust - ideally one-shot - framework, a small value is most desirable here. }
\label{fig:synth_perf}
\end{minipage}
\end{figure}

\twocolumn
\section{Results and Discussion}
\subsection{Synthetic Functions}
A graphical representation of the search, including the optimization progress and model fragility (the variance between runs) is shown in \textbf{Figure \ref{fig:synth_perf}}.  A numerical comparison is shown below in \textbf{Table \ref{tab:synth_res}}.
\begin{table}[htbp]
\centering
    \begin{tabular}{ccccccc}
    \hline
           & \multicolumn{2}{c}{Branin (min)}  & \multicolumn{2}{c}{Camelback (min)}      & \multicolumn{2}{c}{Hartmann (max)}      \\
           & \textit{Mean} & \textit{$\Delta CI$} & \textit{Mean} & \textit{$\Delta CI $}& \textit{Mean}& \textit{$\Delta CI $}\\
           \hline
    AEI    & \textbf{0.406} & \textbf{0.002} & \textbf{-1.000}   &\textbf{ 0.000} & 3.074   & \textbf{0.122} \\
    EI-0.0 & 0.997 & 1.481 & -0.9259   &1.000 & \textbf{3.081}   & 0.439 \\
    EI-0.3 & 0.702 & 1.185 & -0.9499   & 0.816 & 3.0754   & 0.652 \\
\hline
    \end{tabular}
    \label{tab:res_synth}
    \caption{Summary of the results of experiments on synthetic functions.  For Confidence Intervals ($\Delta CI $), smaller values demonstrate reliability over multiple runs. }
\label{tab:synth_res}     
\end{table}
It can be seen that our setting of EI produces superior search capability for the three synthetic functions studied. For all but the 6-dimensional Hartmann function, AEI on average produces the most optimal results, and in all cases it achieves that result with the greatest reliability (smallest value for CI).  This is due, in part to its ability to extract itself from local minima, since in the case of the Branin-Hoo function, the higher means for both settings of EI are due to the algorithm getting stuck in a local minima with a far worse value.   Even in the one case in which AEI did not perform the best - 6-dimensional Hartmann - it can be seen that the average result discovered is extremely close to the best discovered by EI, and for this case AEI demonstrates superior reliability. It is also interesting to observe that in general the AEI search tracks with, or outperforms whichever method is performing best in the early sampling.  Given that this method does not require the tuning parameter of traditional improvement, this can be seen as a validation of the dynamic approach taken here.  
\subsection{Tuning of Machine Learning Algorithms}

As previously described, we test our contextual improvement on two tasks - the tuning of three hyperparameters of a support vector machine for the abalone regression task, and the tuning of five parameters of a 2-hidden-layer multi-layer perceptron for the MNIST classification task.  The results can be seen in \textbf{Table \ref{tab:ml_res}}.

\begin{table}[h!]
\centering
    \begin{tabular}{ccccc}
    \hline
           & \multicolumn{2}{c}{SVM }  & \multicolumn{2}{c}{MLP}     \\
           & \textit{Mean} & \textit{$\Delta CI $} & \textit{Mean} & \textit{$\Delta CI $}\\
           \hline
    AEI    & \textbf{1.940}&	0.006 &	0.253 &	0.086 \\
    EI-0.0 &\textbf{1.940}	&\textbf{0.004}&	0.298&	0.223 \\
    EI-0.3 & \textbf{1.940}	&\textbf{0.004}	&\textbf{0.1938}&	\textbf{0.008} \\
\hline
    \end{tabular}
    \caption{Summary of the results of experiments on the tuning of machine learning algorithms - a support vector machine, and a 2-layer multi-layer perceptron.  For Confidence Intervals ($\Delta CI $), smaller values demonstrate reliability over multiple runs. }
\label{tab:ml_res}     
\end{table}

\begin{figure}[htbp]
\centering
\includegraphics[width=1.0\linewidth]{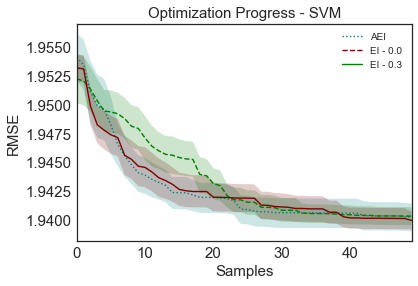}
\caption{Visualization of the search progress for EI with epsilon set to 0.0, and 0.3 and our Adaptive EI, which is based upon contextual improvement for setting hyperparmeters of support vector machines performing the abalone regression experiment. Each experiment is performed 10 times with 3 different randomly selected data points, with confidence intervals are produced by bootstrapping the mean.}
\label{fig:svm_aei}
\end{figure}

For the SVM regression task, it can be seen that methods result in the same results, on average, after 50 epochs, with very little difference in the robustness, although AEI does perform slightly worse.  This could be  indicative of a funnelling shape of the information landscape, in which one basin is both dominant, and wide.  This can be seen in \textbf{Figure \ref{fig:svm_aei}}  This is an ideal case for hyperparameter setting, as the method used does not seem to particularly impact the results although, as can be seen from the other experiments in this study, it is not a typical one.  As the study in the next section clearly shows, however, this could also be due to fortunate choices of which values of $\epsilon$ to study, and the authors argue that in tasks such as hyperparameter searches, which can be critical to the success of tasks further down the pipeline, disconnecting the confidence in the quality of the hyperparameters from the setting of a search hyperparameter such as $\epsilon$ should be considered a significant advantage of this method.

The five-dimensional MLP-classification hyperparameter-setting task was more challenging for AEI, and the best performance was obtained using EI with $\epsilon = 0.3$.  It is worth noting, however, for this task that the performance of $\epsilon = 0.0$ - significantly worse both in search results and in CI - may suggest that the slightly worse performance of AEI is a price worth paying given the potential ramifications of getting the wrong value for $\epsilon$.  Of course, this is said under the assumption that there is no \textit{a priori} knowledge about this value; and if this is not the case then this should be built taken into account when making risk-reward judgements. This is studied and discussed in more detail in the next section.  The authors also recognise the possibility of building this knowledge into the contextual improvement framework, and this is an area under ongoing investigation.

\subsection{Experimental Design}
\begin{table}[htbp!]
\centering
    \begin{tabular}{ccc}
    \hline
           & \multicolumn{2}{c}{Aerofoil }      \\
           & \textit{Mean} & \textit{$\Delta CI $} \\
           \hline
    AEI    & \textbf{255.0327}&	183.7029  \\
    EI-0.0 &234.0355&	195.8036\\
    EI-0.3 & 187.7445	&\textbf{165.3584}	 \\
\hline
    \end{tabular}
    \caption{Summary of the results of experiments on the experimental design of 2D aerofoils, a maximization problem.  For Confidence Intervals ($\Delta CI $) smaller values demonstrate reliability over multiple runs. }
\label{tab:exp_res}     
\end{table}

\begin{figure}[htbp]
\begin{center}
\includegraphics[width=1.0\linewidth]{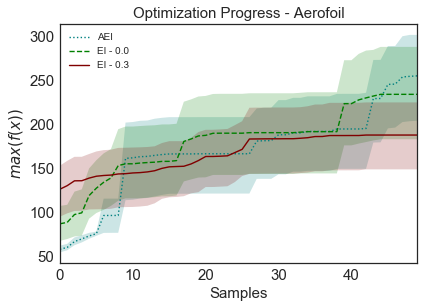}
\end{center}
\caption{Visualization of the search progress for EI with epsilon set to 0.0, and 0.3 and our Adaptive EI, which is based upon contextual improvement. Each experiment is performed 10 times with 3 different randomly selected data points, with confidence intervals produced by bootstrapping the mean.}
\label{fig:aerofoil_search}
\end{figure}

This problem was selected to represent a real-world design problem.  Experimental design is an area in which Bayesian optimization has the potential to provide powerful new capabilities, as traditional design of experiment (DoE) approaches are static and information centric (exploratory), and thus have the potential to be highly inefficient for design tasks.  The performance of our AEI protocol here demonstrates the value of dynamic control of explore/exploit tradeoff.  The results are shown in\textbf{ Table \ref{tab:exp_res}}.  Unlike other problems investigated thus far, the $\epsilon = 0.3$ setting of EI is highly inefficient, producing the worst lift/drag ratios out of the three protocols, although as a result of its exploratory nature it has better reproducibility (lower CI).  As can be seen in \textbf{Figure \ref{fig:aerofoil_search}}, AEI discovers the highest performing aerofoils with more reliability than the next best, the $\epsilon = 0.0$ setting of EI - demonstrating how the method balances the twin goals of performance and reproducibility.   

\subsection{Overall Performance - Sensitivity to hyperparameters}
One way to measure the robustness of AEI is to compare the rankings of the search and CI metrics over the whole range of tasks performed in this study.  Since raw rankings can be misleading (a close second ranks the same as a search in which the gap between methods was much wider) we utilize a normalized ranking using the following method.
\begin{equation}
Z = \frac{s - s\prime}{s_{max} - s_{min}}
\label{eq:norm_rank}
\end{equation}

where $s$ represents the result of a particular strategy, $s\prime$ the result of the best strategy, and $s_{max} - s_{min}$ represent the range of results encountered in the study. 

Calculating the average value for $Z$ across each of the experiments performed in this study is enlightening into the benefit provided by the dynamic control of explore-exploit trade-off (essentially, $\epsilon$).  Our contextual-improvement based strategy (AEI) provides superior results for both search results (i.e. the discoverability of desirable solutions) and  the CI (i.e. the robustness of the search).  Additionally, we can start to estimate the dependency of these metrics upon a good choice of epsilon by comparing the $Z$ scores obtained using $\epsilon = 0$ and $\epsilon = 0.3$.  Comparing the overall Z score (i.e. the combination of search and CI), we see that the difference between the two settings of epsilon is around 78\% of the total value of our dynamic setting (\textbf{Table \ref{tab:norm_rank}}, offering a significant degradation in performance. 

\begin{table}[htbp]
\centering
    \begin{tabular}{cccc}
    \hline
    & \multicolumn{3}{c}{\textit{Z}}      \\
    & Search & \textit{$\Delta CI $}& Overall \\
    \hline
	AEI		&\textbf{0.3910} &	\textbf{0.3278} & \textbf{0.3594}\\
	EI-0.0	&0.7187&	0.7665& 0.7426\\
	EI-0.3		&0.4854	&0.4369& 0.4611\\
	\hline
\end{tabular}
    \caption{Summary of the results of experiments performed during this study using the $Z$ criterion in\textbf{ Equation \ref{eq:norm_rank}}.  Bold indicates the best performing method }
    \label{tab:norm_rank}
\end{table}

\subsection{The importance of a one-shot technique}

It is important to note here that the true \textit{apples to apples} comparison is not really between any one value of $\epsilon$, be it $0.0$, or $0.3$, (or even the difference between these two values) but instead to compare to the CI over a wide range of $\epsilon$ since the correct value cannot be determined \textit{a priori}.  In order to better illustrate this point, we perform two of the tasks described in the paper - the Camelback minimization (a synthetic function) and a 'real world' example of tuning the hyperparameters of an SVM for the abalone regression problem -  over a range of values for $\epsilon$ from 0.0 to 1.0, with a resolution of 0.01 (i.e. 100 values of epsilon).  

\begin{figure}[htbp!]
\begin{subfigure}{.5\textwidth}
\includegraphics[width=\linewidth]{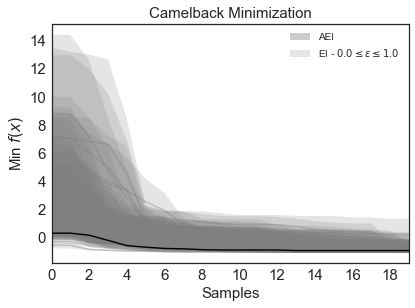}
\caption{•}
\end{subfigure}
\begin{subfigure}{.5\textwidth}
\includegraphics[width=\linewidth]{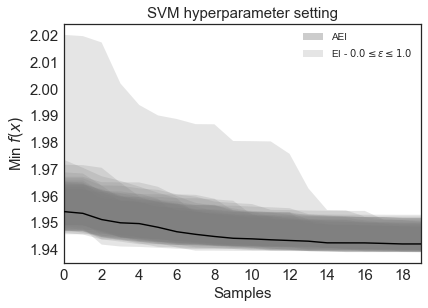}
\caption{•}
\end{subfigure}
\caption{Visualization of the search progress for EI with a set of $\epsilon$  ranging between  0.0, and 1.0 (grey) and our Adaptive EI (black), which is based upon contextual improvement. (a) shows the effect of varying $\epsilon$ for the camelback minimization, whilst (b) shows the effect of varying $\epsilon$ for the SVM hyperparameter search experiment. Each experiment is performed 5 times with 3 different randomly selected data points, with confidence intervals produced by bootstrapping the mean.}
\label{fig:varEPS}
\end{figure}

The additional uncertainty associated with selecting a particular value of $\epsilon$ can be clearly be seen from \textbf{Figure \ref{fig:varEPS}}.  Whilst we can see from the previous experiments that it is possible to find a value of $\epsilon$, which performs as well as AEI, it is hard to know what the best value should be.  \textbf{Figure \ref{fig:varEPS}} shows the potential danger of using a poor value of $\epsilon$, with \textbf{Figure \ref{fig:varEPS}} (b) showing clearly the potential danger of choosing a bad value for epsilon when samples are low.  In the typical Bayesian optimization setting, this is particularly important as there may be very little sampling as performing a ground truth evaluation can result in a significant cost, either financial or computational and thus a method which minimizes this risk has significant benefits.  Additionally, since many decision making exercises are coming to increasingly rely on deterministic (i.e. not Bayesian), but highly scalable machine learning models, the potential consequences of not locating a good set of hyperparameters can be significant.  'One shot' methods such as AEI afford the user a larger degree of confidence that the search has located a good set of parameters without the need to evaluate multiple search settings (such as would be required with $\epsilon-EI$).  

An approximation to the risk reward trade-off can be performed visually using \textbf{Figure \ref{fig:varEPS}}.  Experiments in which the gamble failed to pay dividends (i.e. the performance of using a constant $\epsilon$ is worse than AEI) are represented as the shaded area above the black trace.  This can be thought of as the situations in which AEI outperforms a static $\epsilon$ model.   It can be seen that or both tasks evaluated there is a large density of experiments which fall into this `loss' zone, especially when small number of samples have been drawn.  For an idea of the magnitude of the risk, you can compare the areas shaded grey above and below the black trace.  Again, the expectation, given a random selection of $\epsilon$ is significantly in the `loss' with this result being more pronounced at low number of samples. 

\section{Conclusion}

We present a simple, yet effective adaptation to the traditional formulation of improvement, which we call contextual improvement.  This allows a Bayesian optimization protocol to dynamically schedule the trade-off between explore and exploit, resulting in a more efficient data-collection strategy.  This is of critical importance in Bayesian optimization, which is typically used to optimize functions where each evaluation is expensive to acquire.  We have demonstrated that EI based upon contextual improvement outperforms EI using traditional improvement, and improvement with a margin in a range of tasks from synthetic functions to real-world tasks, such as experimental design of 2-D NACA aerofoils and the tuning of machine learning algorithms.  We also note that our proposed contextual improvement results in settings of expected improvement which are significantly more robust to the random seed data, which is a highly desirable property since this allows the use of minimal seed data sets.  In traditional Bayesian optimization settings, where each data point is expensive to acquire, this can result in significant savings in costs, both in time and financial outlay.  

\section{Acknowledgements}

The authors thank Dr Kirk Jordan for helpful discussions.

\end{document}